\documentclass[runningheads]{llncs}

\usepackage{eccv}

\usepackage{eccvabbrv}

\usepackage{graphicx}
\usepackage{booktabs}
\usepackage{multirow}

\usepackage[colorlinks=true, urlcolor=magenta]{hyperref}
 
\usepackage[accsupp]{axessibility}  

\usepackage{hyperref}

\usepackage{orcidlink}

\begin{document}

\renewcommand{\thefootnote}{\fnsymbol{footnote}}

\title{MOFA-Video: Controllable Image Animation via Generative Motion Field Adaptions in Frozen Image-to-Video Diffusion Model} 

\titlerunning{MOFA-Video}

\author{Muyao Niu\inst{1} \and
Xiaodong Cun\inst{2,\ensuremath{\star}} \and
Xintao Wang\inst{2} \and
Yong Zhang\inst{2} \and \\
Ying Shan\inst{2} \and
Yinqiang Zheng\inst{1,\ensuremath{\star}}
}

\authorrunning{M. Niu et al.}

\institute{
The University of Tokyo \\
\email{muyao.niu@gmail.com}, \email{yqzheng@ai.u-tokyo.ac.jp} \and
Tencent AI Lab \\
\email{vinthony@gmail.com}, \email{xintaowang@tencent.com}, \\ \email{zhangyong201303@gmail.com}, \email{yingsshan@tencent.com} \\ \vspace{1mm}
\url{https://myniuuu.github.io/MOFA_Video/} 
}

\maketitle

\footnotetext[1]{Corresponding authors}

\renewcommand{\thefootnote}{\arabic{footnote}}

\begin{figure}
\centering
  \includegraphics[width=1\textwidth]{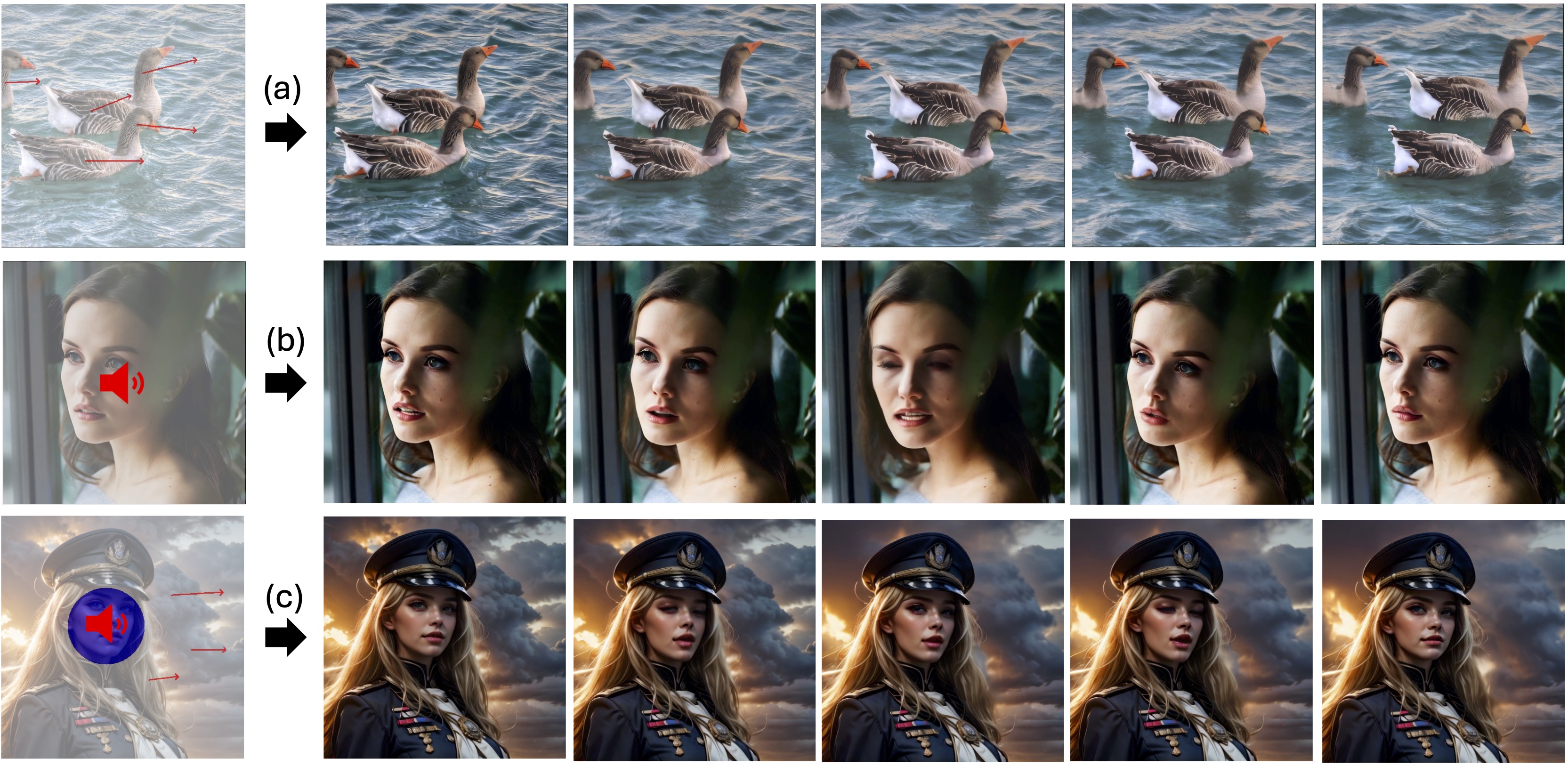}
  \vspace{-2em}
  \caption{We present MOFA-Video for controllable image animation. We train MOFA-Adapters for (a)~manual trajectories animation, (b)~facial landmarks sequences animation~(SadTalker~\cite{zhang2023sadtalker} is used for audio to landmark generation). These two adaptors can be combined in a zero-shot manner for (c)~the animation from both trajectories and human landmarks without retraining.}
  \vspace{-3em}
  \label{fig:teaser}
\end{figure}

\begin{abstract}
We present MOFA-Video, an advanced controllable image animation method that generates video from the given image using various additional controllable signals~(such as human landmarks reference, manual trajectories, and another even provided video) or their combinations. This is different from previous methods which only can work on a specific motion domain or show weak control abilities with diffusion prior. To achieve our goal, we design several domain-aware motion field adapters~(\ie, MOFA-Adapters) to control the generated motions in the video generation pipeline. For MOFA-Adapters, we consider the temporal motion consistency of the video and generate the dense motion flow from the given sparse control conditions first, and then, the multi-scale features of the given image are wrapped as a guided feature for stable video diffusion generation. We naively train two motion adapters for the manual trajectories and the human landmarks individually since they both contain sparse information about the control. After training, the MOFA-Adapters in different domains can also work together for more controllable video generation. Codes available: \url{https://github.com/MyNiuuu/MOFA-Video}

\end{abstract}
\section{Introduction}
\label{sec:intro}

Bringing images to life is considered magic in the old days. In the traditional Chinese story ``The Magic Brush Ma Liang'', the author imagines a magic pen that can directly draw the living photos. Coincidentally, the story of Harry Potter creates a world where the dead ancestors are living in the wall paintings. Besides fiction, the exploration to make it come true has never stopped. In 1878, Muybridge presented a well-known experiment called ``The Horse in Motion'', which shows a series of successive pictures of the running horse consecutively that can be considered a video. 

%
With the development of digital devices, current methods attempt to animate the image using computer vision algorithms~\cite{mahapatra2021controllable, Siarohin_2021_CVPR,  Weng_2019_CVPR, Karras_2023_ICCV, Blattmann_2021_ICCV, ni2023conditional, li2023generative, Holynski_2021_CVPR, rottshaham2019singan, Xiong_2018_CVPR, zhang2023sadtalker, wang2022latent, Dorkenwald_2021_CVPR}. However, it faces several limitations. On the one hand, these methods often focus on limited categories of animated objects such as fluids~\cite{Holynski_2021_CVPR, mahapatra2021controllable, mahapatra2023synthesizing}, human hairs~\cite{hairAni}, and human body/face~\cite{zhang2023sadtalker, Bertiche_2023_CVPR,Blattmann_2021_CVPR,Karras_2023_ICCV,Siarohin_2021_CVPR,Weng_2019_CVPR,wang2022latent,Blattmann_2021_ICCV,Dorkenwald_2021_CVPR,ni2023conditional}. Thanks to the domain knowledge of each specific type, these methods often have fully controllable abilities of the scenes. \eg, SadTalker~\cite{zhang2023sadtalker} can produce the accurate human face animation by audio and the given face. Text2Cinemagraph~\cite{mahapatra2023synthesizing} produces the natural animation of water using text description. For control abilities, these methods usually follow the rule of learning the video via self-supervised decomposition and then animating via new driving signals. However, due to the limitation of natural animation prior, these methods fail in the general image space because of the diversity of the general domain knowledge.

Unlike previous in-domain image animation, current diffusion-based Image-to-Video~(I2V) methods learn to generate the video from the image in an end-to-end fashion. Thanks to the large-scale generative prior of the text-to-image model, \ie, Stable Diffusion~\cite{sd}, these methods~\cite{svd, xing2023dynamicrafter, chen2023videocrafter1, genmo, gen2} have proved the possibility of open-domain image animation. However, their generated contents might be different from the given image~\cite{gen2, genmo, chen2023videocrafter1, xing2023dynamicrafter}, and often generate simple motions by text descriptions~\cite{xing2023dynamicrafter, gen2, genmo} or just simple idle animation~\cite{svd}. These drawbacks limit their applications for real-world image animation tasks, where users often need to create more controllable videos as in previous in-domain image animation algorithms.

Taking advantage of both in-domain image animations and image-to-video generations, we are curious: \textit{is there a general image animation framework that supports meticulous control abilities over in-the-wild images?} We then find that all the animations can be formulated by the motion propagation of the sparse key-points~(or key-frame) as the control handle.

To this end, we present MOFA-Video to add the additional different motion control abilities to the generatic video diffusion model~(Stable Video Diffusion~\cite{svd} in our case), inspired by ControlNet~\cite{controlnet}. In detail, to animate an input image into a video according to the sparse control signals from multiple domains, we design a novel MOFA-Adapter that serves as an additional adapter on the pre-trained video diffusion model so that the motions of the video can be controlled. 
Different from the previous ControlNet-like Adapter~\cite{xing2023make, controlvideo} for video generation, the proposed MOFA-Adapter models the frame-wise motion explicitly. In detail, we first utilize the given sparse motion hint to generate the dense motion fields using a sparse-to-dense motion generation network, then, we warp the multi-scale features of the first frame as the conditional features for the diffusion generation process. This sparse-to-dense motion generation provides a good balance between the provided motion guidance and the generation process, providing high-quality animation results with good temporal consistency. We also think about the problems of there containing multiple motion domains. Thus, we train multiple MOFA-Adapters by considering these tasks as sparse control point generation problems, including open-world manual trajectories, human facial animations, \etc. In addition, since the parameters of the video diffusion model are fixed, we can jointly perform motion control abilities across multiple domains, \eg, the human face and background objects and the camera movement. We give more detailed applications and examples in the experiments. 

The contribution of this paper can be summarized as:
\begin{itemize}
    \item We propose a novel unified framework for controllable image animation in Stable Video Diffusion~(SVD).
    \item We design a novel network structure, \ie, MOFA-Adapter, which utilizes the explicit sparse motion hint for warping and generation. 
    \item Detailed experiments and ablation show the advantage of the proposed method over current ones. 
\end{itemize}

\section{Related Work}

\subsection{Controllable Image Animation}
Image animation has a long history in computer vision and graphics. Since it is a very ill-posed problem, previous work only focuses on the specific domain. \eg, previous methods to generate the video from the image in an unsupervised and stochastic manner~\cite{xiong2018learning, li2018flow, Holynski_2021_CVPR, logacheva2020deeplandscape} using generative model, \eg, generative adversarial networks~(GAN). These methods do not provide the control handle of the generation, limiting its applications on real-world applications. On the other hand, controlling the motion of the generation is also difficult for in-the-wild images. To this end, recent works aim to control the fluids~\cite{mahapatra2021controllable, mahapatra2023synthesizing}, human pose/face~\cite{Blattmann_2021_CVPR,Karras_2023_ICCV,Siarohin_2021_CVPR,Weng_2019_CVPR,wang2022latent,Blattmann_2021_ICCV,Dorkenwald_2021_CVPR,ni2023conditional} only, other than the general scenes. Different from these methods, we propose a novel and unified framework to model the motions from different domains and make it work singly and together in a pre-trained Video Diffusion Model.

\subsection{Image-to-Video Diffusion Models.}
In the realm of downstream tasks utilizing VDMs for video-related applications, there exists a category of work known as Image-to-Video Diffusion Models (I2Vs)~\cite{gen2, genmo, pikalabs, xing2023dynamicrafter,i2vgenxl, svd, Zhang_2024_CVPR}. 
However, current I2V models only generate the video from the given image and aim to control the motions from the text description, we argue it is somewhat hard for text-based motion generation since they contain limited motion priors. Some related methods also try to control the generation of the motion. \eg, DragNUWA~\cite{yin2023dragnuwa} model the trajectory of the generation via adaptive normalization, which shows weak spatial correction. MotionCtrl~\cite{wang2023motionctrl} tries to control the object and camera motions of the T2V model, however, it is hard for the generation process since there is no world coordinate system for text-to-video generation. The most related controllable image animation is the concurrent work MotionI2V~\cite{shi2024motion}. However, it focuses only on the natural motion of the objects. which is hard to fully control the motions from other domains, \eg, the human face or body. Differently, we aim to create a method which is built on a pre-trained base model, \ie, SVD, and also enjoys the motion hints from different domains using different adapters.

\section{Method}
\label{sec:method}

\begin{figure}[t]
\centering
\includegraphics[width=\linewidth]{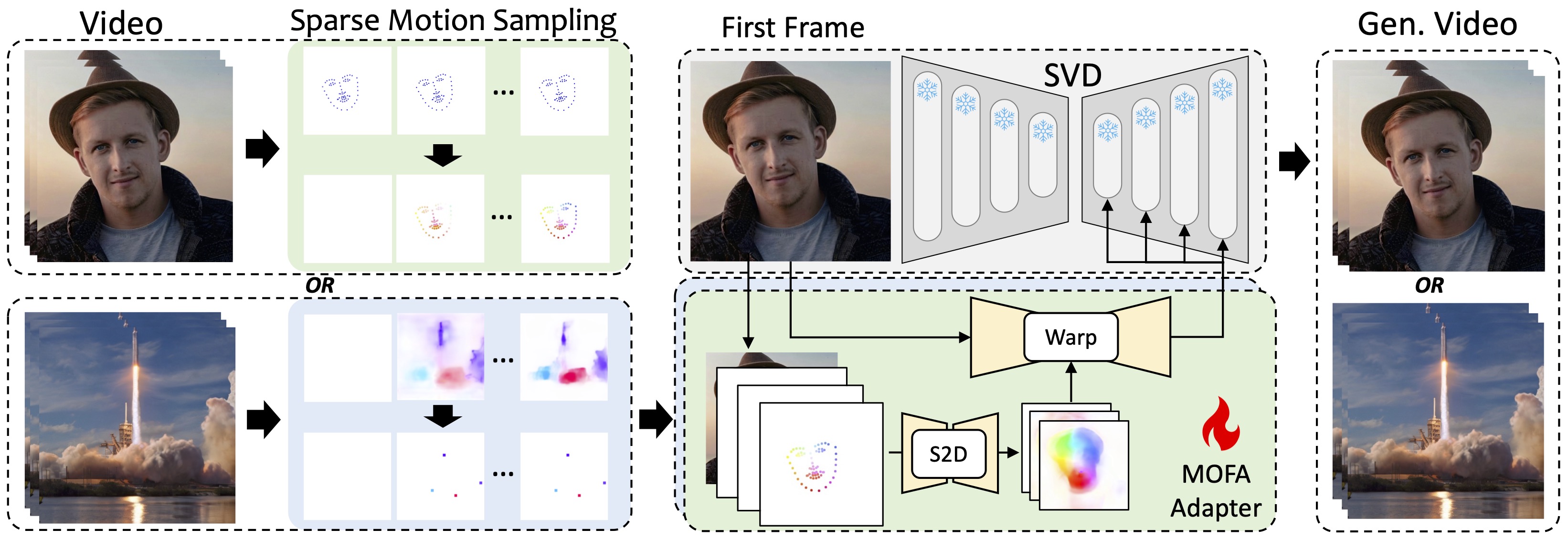}
\vspace{-2em}
\caption{
\textbf{\textit{Overview of MOFA-Video.}} We design MOFA-Adadpters for adapting the motions from different domains with a unified structure on the frozen Video Diffusion Model. It generates the video from a single image and the corresponding sparse motion hints. For training, we generate the sparse motion hints through sparse motion sampling and then train different MOFA-Adapters to generate video via pre-trained SVD~\cite{svd}.
}
\label{fig:overview}
\vspace{-1em}
\end{figure}

We aim to generate videos from the given reference image and additional motion control signals in multiple motion domains~(\eg, hand-crafted trajectories, human landmark sequences, dense motion flows, \etc.) in a unified framework, so they can share a unified network structure and jointly work together like Multi-ControlNet~\cite{controlnet}. 
To achieve this goal, as shown in Fig.~\ref{fig:overview}, we design a generative motion field adapter~(MOFA-Adapter) that can accept sparse motion control signals as the condition and produce detailed control abilities for a frozen Stable Video Diffusion~\cite{svd} Model. We train the proposed MOFA-Adapter in two different motion domains individually and provide various applications based on each model and their combinations.

In the following, we first introduce the structure of the proposed MOFA-Adapter in Sec.~\ref{sec:mofa-adapter}. Then, we give details of how we train the domain-aware MOFA-Adapter for the video diffusion model in Sec.~\ref{sec:svd_training}. Finally, in Sec.~\ref{sec:inference}, we give the inference details of the proposed method and various additional applications.

\begin{figure}[t]
\centering
\includegraphics[width=\columnwidth]{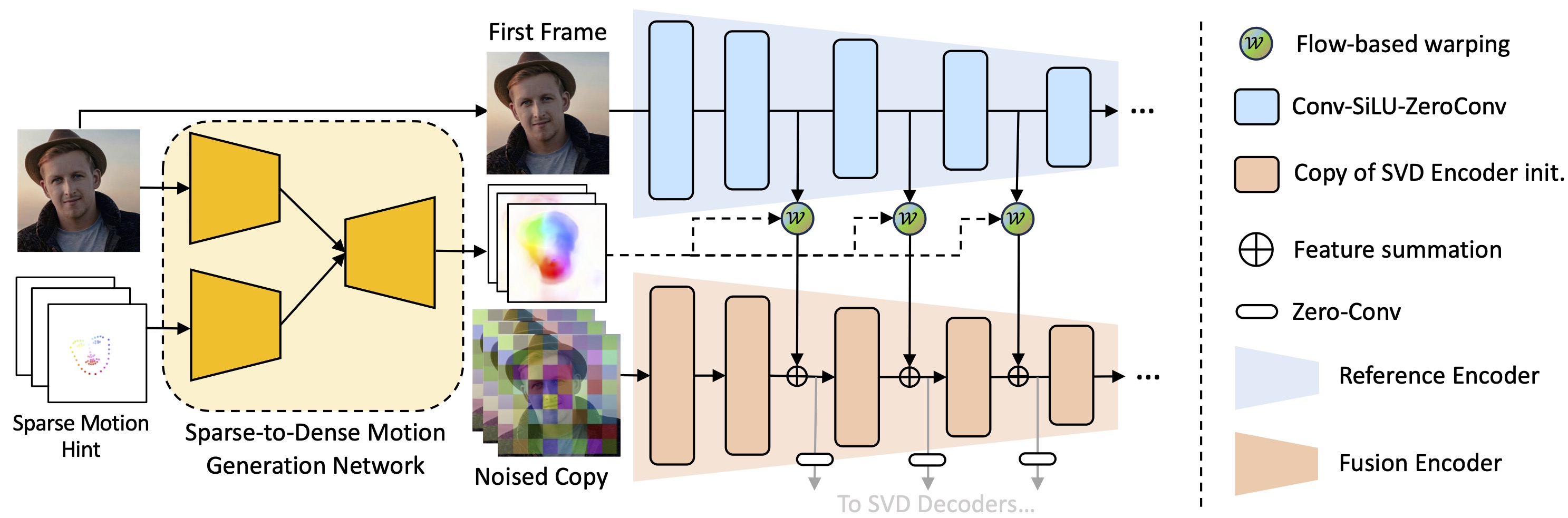}
\vspace{-2em}
\caption{
\textbf{\textit{Detailed Structure of MOFA-Adapter.}} It contains an S2D Network that accepts the motion hints and produces a dense motion field of the video. A reference encoder that extracts multi-scale features from the source image. A training-able copy of the SVD encoder, which initializes the weights from SVD and serves as the final spatial-temporal feature merging for generation guidance. 
}
\label{fig:mofa-adapter}
\vspace{-2em}
\end{figure}

\subsection{Generative Motion Field Adapter~(MOFA-Adapter)}
\label{sec:mofa-adapter}

Given the sparse motion hints~(\eg, hand-crafted trajectories, human pose sequential trajectories), we design a new adapter structure, \ie, MOFA-Adapter, for the pre-trained frozen Video Diffusion Model, so that the generated video of different subjects can be controlled individually or jointly in the generation process. Inspired by ControlNet~\cite{controlnet}, we consider this network as an additional motion control signal to the frozen denoising UNet of the video diffusion model. Below, we give the detailed network structure.

As shown in Fig.~\ref{fig:mofa-adapter}, the proposed adapter is based on a reference encoder, a sparse-to-dense~(S2D) motion generator for sampled motion hints, and the feature fusion encoder to add the warped feature back to the pre-trained video diffusion model. 
In detail, the reference image encoder is a multi-scale convolutional feature encoder, which extracts the multi-scale features of the first frame for warping, with each stage built by Conv-SiLU-ZeroConv~\cite{controlnet} as the basic block. 
For the sparse-to-dense motion generator, we use the same network structure as CMP~\cite{zhan2019self} for adaption. 
This network is also a convolutional neural network which accepts the first frame image and the sparse hint of the motion, and produces the dense motion fields. 
More details of the network structure can be founded in the original paper~\cite{zhan2019self}. 
When the dense motion field is generated, we warp the referenced features and then add them to the feature map of the corresponding levels of a copied SVD encoder, which is then added to the feature space of the decoder of the pre-trained SVD similar to ControlNet~\cite{controlnet}.

\subsection{Training MOFA-Adapters on Stable Video Diffusion}
\label{sec:svd_training}

We use Stable Video Diffusion~\cite{svd} as our base image-to-video diffusion model which accepts the image as input and generates video with idle animations. It is a latent diffusion model~\cite{sd} which firstly compresses the reference image into the latent space using a pre-trained auto-encoder, and then, the video is generated via the sampled Gaussian noise, the condition image, and the diffusion process~\cite{ddpm}.

Fig.~\ref{fig:overview} illustrates the training pipeline of our whole framework. Given a video clip $ V \in \mathbb{R}^{L\times 3\times H\times W}$ in $L$ frames, we first extract the sparse motion vectors which serve as input to the S2D network.
\eg, for open-domain, we handle the motion hint as the sparse motion vectors sampled from the extracted dense optical flow. As for human motion, we generate the motion hints from structural key-points, \eg, facial landmarks. In the following, we give the details of each specific type:

\noindent \textbf{Sparse Motion Vectors from Dense Optical Flow.} 
By considering the dense optical flow as a general motion representation between video frames, we first utilize Unimatch~\cite{xu2023unifying} to extract the forward flow as $F \in \mathbb{R}^{(L-1)\times 2\times H\times W}$, where $F_i \in \mathbb{R}^{2\times H\times W}$ represents the optical flow from the $0$-th to the $(i+1)$-th frames. Based on the flow sequence $F$, we sample $n$ spatial points for each frame $F_i$ using the watershed sampling strategy~\cite{zhan2019self}. Specifically, we first obtain a sparse mask $M^s \in \mathbb{R}^{H\times W}$, where the value for the sampled spatial points is set to 1, and other points are set to 0. We then calculate the sparse motion vectors $F^s \in \mathbb{R}^{(L-1)\times 2\times H\times W}$ as:
\begin{align}
    F^s_{:,:,i,j} = \begin{cases}
                      F_{:,:,i,j} & \text{if } M^s_{i,j} = 1, \\
                      0  & \text{if } M^s_{i,j} = 0.
                    \end{cases}
\end{align}

\noindent \textbf{Sparse Motion Vectors from Structural Human Key-Points.} 
Different from natural motion fields, human key-points provide concise while semantically meaningful representations. In our approach, we consider the movement of a group of key-points as a special case of the sparse motion vectors mentioned earlier. This unified representation simplifies our framework and allows us to share the mutual prior information of the S2D model.
Specifically, given a series of 2D facial landmarks $P \in \mathbb{R}^{L \times K \times 2}$ extracted from an $L$-frames portrait video, we consider the motion difference between the landmarks of reference~(first) frame $\hat{P}$ and $P$, calculating point-wise sparse flow $F^s$ via:
\begin{equation}
\label{equ:keypoint}
\begin{aligned}
    &F^s[l-1, :, \hat{P}[k,0], \hat{P}[k,1]] = P[l,k,:] - \hat{P}[k, :],
\end{aligned}
\end{equation}
where $\ l \in \{1, 2, ..., L-1\}. \ k \in \{0, 1, ..., K - 1\}$. This motion mapping enables us to incorporate key-point information into our framework effectively as Fig.~\ref{fig:overview}.

After unifying the motion guidance from different domains, following the training process of stable video diffusion~\cite{svd}, our model is expected to reconstruct the video clips $V$ using the first frame $I$ and the sampled sparse motion vector $F^{s}$. Formally, the diffusion-based method first encodes video $V$ into the latent space through $\mathcal{V} = \mathcal{E}(V)$, then progressively adds noise $\{\epsilon_t\}^{T-1}_{t=0}$ to $\mathcal{V}$ to produce $\{\mathcal{V}_t\}^{T-1}_{t=0}$, where $t$ represents the time step of the added noise. With the pre-trained Stable Video Diffusion $\mathcal{S}$ and the MOFA-Adapter $\mathcal{M}$, we optimize the weights of the MOFA-Adapter $\theta_{\mathcal{M}}$ via :
\begin{equation}
   \mathcal{L} = || \mathcal{S}(\mathcal{V}_t, t, \mathcal{M}(\mathcal{V}_t, t, I, F^{s}; \theta_{\mathcal{M}})) - \mathcal{V} ||^2 ,
\end{equation}
where $\mathcal{L}$ is the overall learning objective. After that, the video diffusion model is run $T$ time steps to recover the clear latent from pure sampled Gaussian noise. Finally, the recovered latent is decoded via the pre-trained auto-encoder to the sequence of the image as the produced video.

\subsection{Inferences}
\label{sec:inference}

After training, our method can generate video from a single image and the given control signals, \eg, handcrafted trajectories, facial key-points, \etc. 
In the following parts, we introduce several inference pipelines and applications of our method, including hand-crafted trajectory-based animation, motion brush masking, and portrait animation with key-points, \etc.

\begin{figure}[t]
\centering
\includegraphics[width=\linewidth]{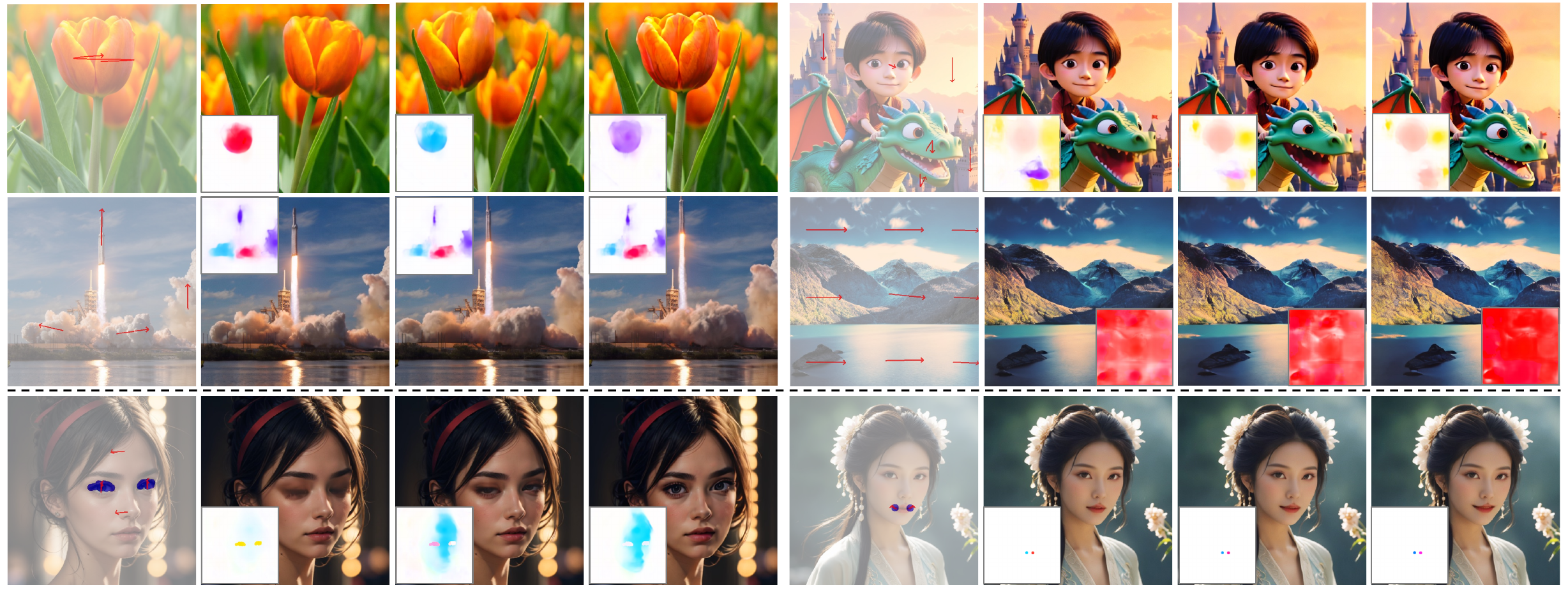}
\vspace{-2em}
\caption{
\textbf{\textit{Trajectory-based Animation.}}
Image animation results from different trajectories. Results below the dashes are the fine-grained results using motion brushes. Intermediate optical flow results are also visualized.
}
\label{fig:naive_drags}
\vspace{-15pt}
\end{figure}

\noindent \textbf{Hand-Crafted Trajectories for Objects/Camera Movement.} 
Given an input image $ I \in \mathbb{R}^{H \times W \times 3}$, the users can draw multiple motion trajectories on the image, each of which can be represented as $P$ joint points $\mathcal{T} \in \mathbb{R}^{P \times 2} = [(x_0, y_0), (x_1, y_1), ..., (x_{P-1}, y_{P-1})]$. Take one trajectory for example, we bicubicly interpolate this trajectory to $L$ points, \ie, $\hat{\mathcal{T}} \in \mathbb{R}^{L \times 2}$ and calculate the sparse motion hints $F^s \in \mathbb{R}^{(L-1) \times 2 \times H \times W}$ between the frames as:
\begin{equation}
\begin{aligned}
    F^{s}[l-1, :, \hat{\mathcal{T}}[0,0], &\hat{\mathcal{T}}[0,1]] = \hat{\mathcal{T}}[l, :] - \hat{\mathcal{T}}[0, :], \\
    \text{where} \ l &\in \{1, 2, ..., L-1\}.
\end{aligned}
\end{equation}
As a result, we can create several trajectories to indicate the motions of the specific object individually. As shown in Fig.~\ref{fig:naive_drags}, we can handle trajectories in both linear and non-linear fashion since our model is able to handle both these two types of motion via optical flows. In addition, by adding relatively dense guidance on the whole images, our method also has the capability to control the motions of the entire scene to perform as ``camera motion''.

\noindent \textbf{Regional Motion Brushes.} 
Besides simple hand-craft trajectories, we can generate the regional motion animation based on the trajectory model. This feature is different from current implicitly drag-based methods~(DragNUWA~\cite{yin2023dragnuwa}), where the animated region is determined by the model itself instead of the user. Specifically, the user could additionally provide a binary motion mask $M \in \mathbb{R}^{H \times W}$ to control the animation region. 
Given this mask, we divide the trajectory set $\mathcal{T}$ into two groups: $\mathcal{T}_{\text{in}}$ and $\mathcal{T}_{\text{out}}$. For each trajectory, we distribute it to $\hat{\mathcal{T}}_{\text{in}}$ if its start points lie in the mask, otherwise to $\hat{\mathcal{T}}_{\text{out}}$. We then respectively obtain the dense optical flow $F_{in}$ and $F_{out}$ via the S2D network, and mask out all the values that are located outside the mask. We then obtain the final dense optical flow by combining the masked $F_{in}$ and $F_{out}$. We demonstrate the corresponding results in Fig.~\ref{fig:naive_drags}.

\begin{figure}[t]
\centering
\includegraphics[width=\linewidth]{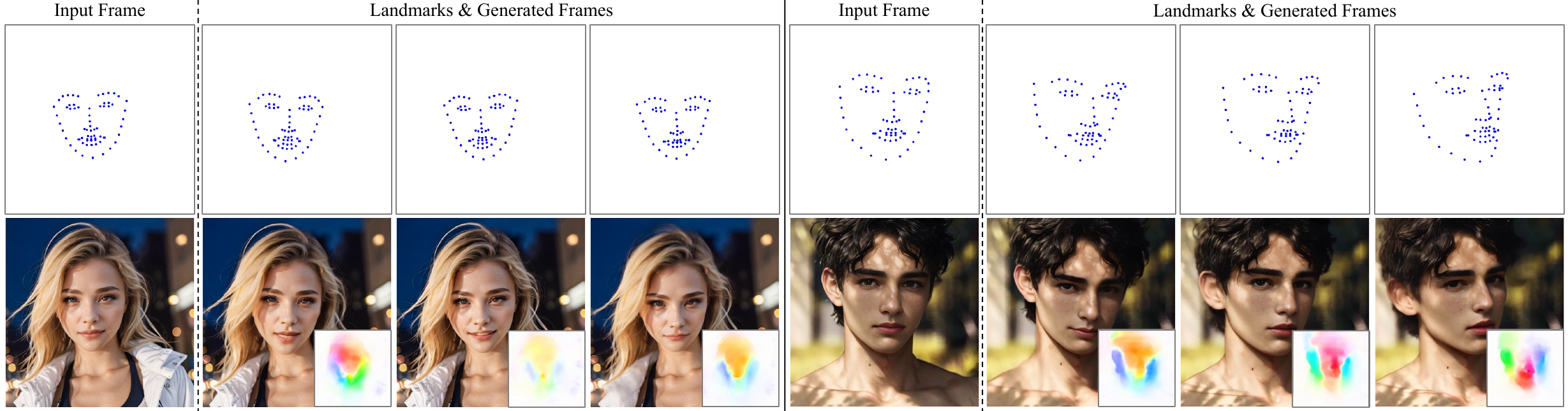}
\vspace{-2em}
\caption{
\textbf{\textit{Facial Landmarks based Animation.}}
We produce facial landmarks from the driven audio using SadTalker~\cite{zhang2023sadtalker}. Then, the portrait animation can be produced by the facial MOFA-Adapter. Intermediate optical flow results are also visualized.
}
\label{fig:keypoint}
\vspace{-10pt}
\end{figure}

\noindent \textbf{Portrait Image Animation.}
Besides natural object movement, we also consider the motion domain of the human, specifically the simple face movement, which is more detailed and structural.
For inference animation, we first use an audio-driven talking face generation method, \ie, SadTalker~\cite{zhang2023sadtalker}, to produce facial landmarks from a single image, then, as introduced in Sec.~\ref{sec:mofa-adapter}, we generate the sparse motion hints and use the trained MOFA-Adapter for motion generation.
Fig.~\ref{fig:keypoint} demonstrates image animation results via audio-driven facial key-points generation~\cite{zhang2023sadtalker} and a single image. Notice that, since we use facial landmarks as the intermediate representation, we can also drive the animation with another facial video or manual facial blendshapes animation.

\begin{figure}[t]
\centering
\includegraphics[width=\linewidth]{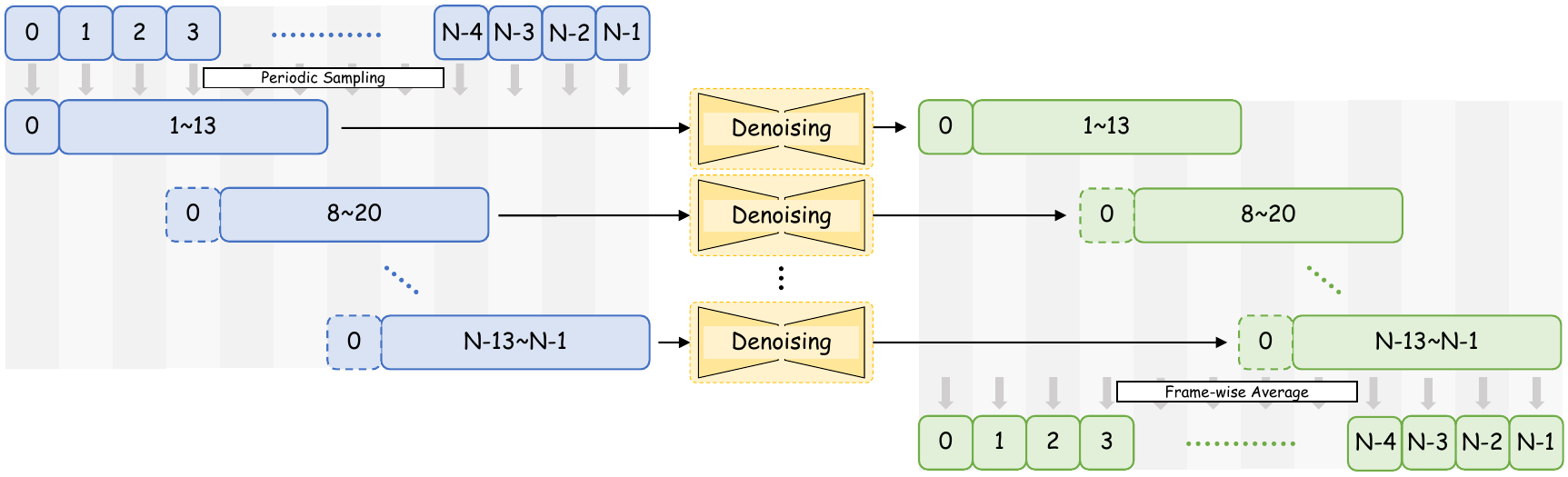}
\vspace{-2em}
\caption{
\textit{\textbf{Periodic Sampling for Longer Animation}}. The first frame~($0$) is used as the condition and we produce latent by frame-wise average over sequential 7 frames.
}
\label{fig:periodic}
\end{figure}

\noindent \textbf{Longer Animation.}
Despite the significant generation ability provided by the SVD, the frame length of its generated video is extremely limited to the specific frames~(\eg, 14 in our case). The computational complexity of temporal attention scales quadratically with the number of frames, making it difficult to directly generate ultra-long videos. On the other hand, ensuring consistency becomes more challenging with the increase in the number of frames. To solve this limitation, we propose the periodic sampling strategy to resolve the frame number issue and generate longer videos, inspired by Gen-L-Video~\cite {gen-l-video}. As illustrated in Fig.~\ref{fig:periodic}, in each diffusion step, we periodically sample 14 frames as one group and denoise them using our SVD model. Each latent group has an overlapping of 7 frames to provide temporal smoothness. After all the groups are sampled, we average the predicted noise for each frame and obtain the final latent for this diffusion step.

\begin{figure}[t]
\centering
\includegraphics[width=\linewidth]{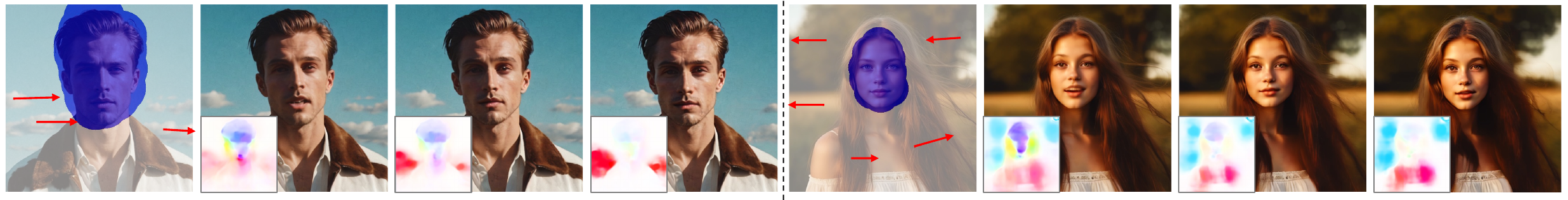}
\vspace{-2em}
\caption{\textbf{\textit{Animation via Multi MOFA-Adapters.}} We show the combination of the hand-craft trajectories and audio-driven facial landmarks-based MOFA-adapters for more complex animations. 
}
\label{fig:mixed_animation}
\vspace{-15pt}
\end{figure}

\noindent \textbf{Plugin Combinations.} Our careful design enables us to animate the given image by simultaneously using different control signals from multiple modalities without any retraining. This is achieved by combining multiple MOFA-Adapters for the animation similar to Multi-ControlNet~\cite{controlnet}. In Fig.~\ref{fig:mixed_animation}, we show an example of controlling the facial expression of one portrait while animating other objects such as backgrounds via hand-craft trajectories. Please refer to the supplementary materials for more experiment results and technical details.

\if

\fi

\section{Experiment}

\noindent \textbf{Implementation Details.}
For training, we train the proposed trajectories-based MOFA-Adapter on WebVid-10M~\cite{webvid} using 8 NVIDIA A100 GPU. The base model is Stable Video Diffusion~\cite{svd}. We use AdamW~\cite{loshchilov2017decoupled} as an optimizer. During training, we randomly sample 14 video frames with a stride of 4. The learning rate is set to $ 2 \times 10 ^{-5} $ with a resolution of $256 \times 256$. We first train the model as a flow-based reconstruction model by removing the S2D motion generator and directly taking the first frame together with the estimated optical flow from Unimatch~\cite{xu2023unifying}. After that, we add the S2D network and finetune the whole adapter.
For portrait-based MOFA-Adapter, we train our model on a self-collected human video dataset. The pre-trained DWPose~\cite{dwpose} is used to extract the facial landmark and other training settings are the same as the trajectories-based model. Please refer to the supplementary for more details. 

\noindent \textbf{Evaluation Metrics.}
For the trajectory-based MOFA-Adapter, we use 1000 samples from the test set of WebVid and respectively calculate LPIPS~\cite{zhang2018perceptual}, FID~\cite{heusel2017gans}, and FVD~\cite{digan}. We also calculate the cosine similarity of the CLIP~\cite{radford2021learning} embedding of the consecutive generated frames to evaluate the frame consistency of the results. To evaluate the keypoint-based MOFA-Adapter, we choose 40 generated video examples, each of which contains 196 frames, and calculate the following two metrics inspired by previous work~\cite{zhang2023sadtalker}: 1) Cumulative Probability Blur Detection (CPBD)~\cite{narvekar2011no} to evaluate the sharpness of the generated frames, and 2) identity embedding from ArcFace~\cite{deng2019arcface} between the source images and the generated frames to evaluate the fidelity of the generated results. 
Apart from the quantitative metrics, we also conduct user studies to perceptually evaluate the result of different methods.

\begin{figure}[t]
\centering
\includegraphics[width=\linewidth]{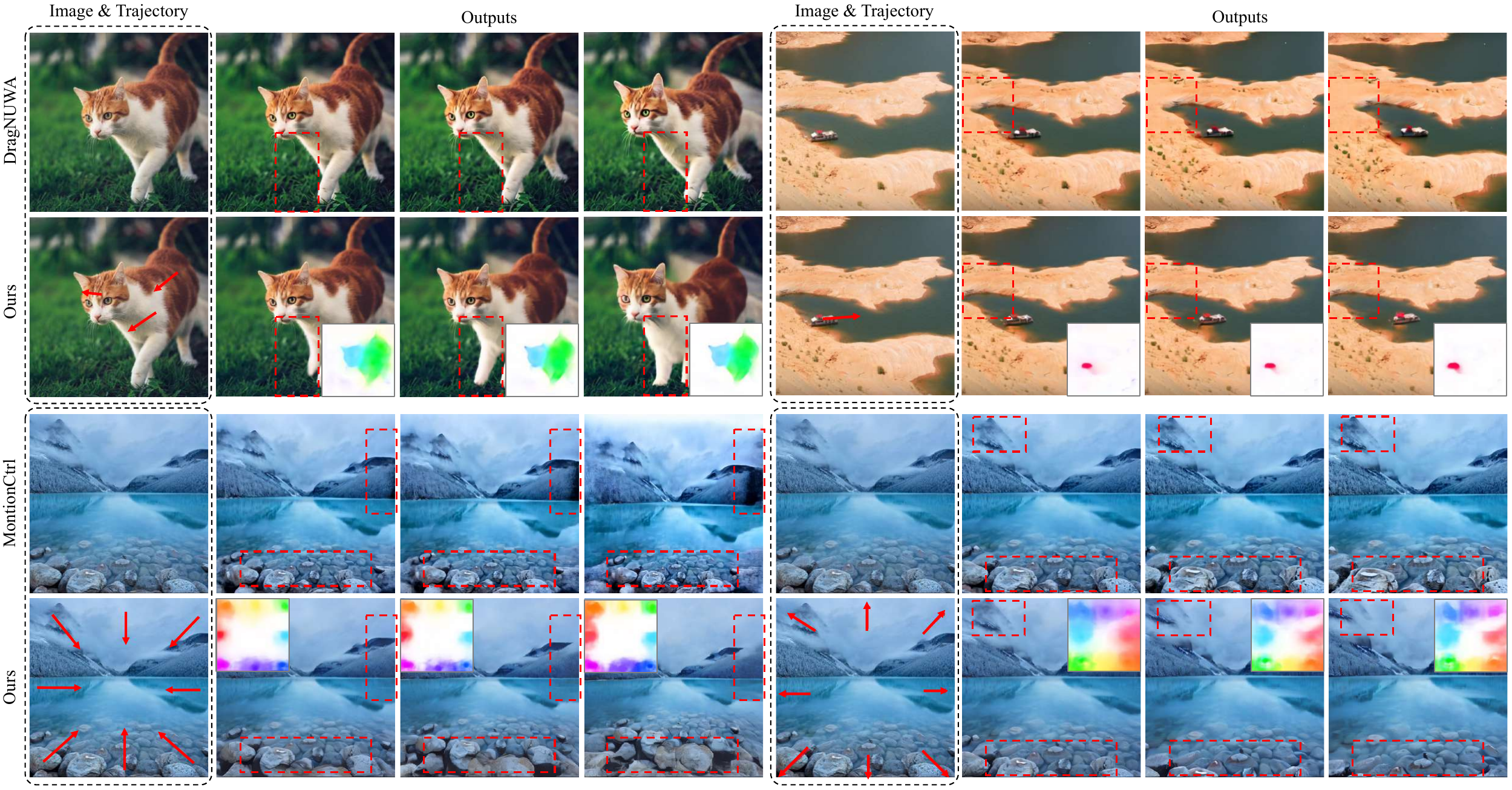}
\vspace{-2em}
\caption{
Comparing trajectory-based control with DragNUWA~\cite{yin2023dragnuwa} and Camera control with MotionCtrl~\cite{wang2023motionctrl} using proposed trajectory-based MOFA-Adapter.
}
\label{fig:comparison}
\vspace{-10pt}
\end{figure}
\begin{table*}[t] 
\centering
\begin{minipage}[b]{0.59\textwidth}
\centering
\resizebox{\textwidth}{!}{%
\begin{tabular}{l|cccc|cc}
\toprule
\multirow{2}{*}{Methods} & \multicolumn{4}{c|}{Metrics} & \multicolumn{2}{c}{User Preference} \\
\cline{2-5} \cline{6-7}
 & \small{Fra.Con.$\uparrow$} & \small{LPIPS$\downarrow$} & \small{FID$\downarrow$} & \small{FVD$\downarrow$} & \small{Ctrl.Pre.$\uparrow$} & \small{Vis.Qua.$\uparrow$} \\
\hline
DragNUWA & 0.9302 & 0.2705 & 19.66 & 91.38 & 2.76 & 3.18 \\
\hline
Ours & \textbf{0.9390} & \textbf{0.2274} & \textbf{16.82} & \textbf{86.76} & \textbf{3.58} & \textbf{3.42} \\
\bottomrule
\end{tabular}
}
\caption{Quantitative comparison and user study results for trajectory-based image animation.}
\label{tab:metric_traj}
\end{minipage}
\hfill 
\begin{minipage}[b]{0.40\textwidth}
\centering
\resizebox{\textwidth}{!}{%
\begin{tabular}{l|cc|ccc}
\toprule
\multirow{2}{*}{Methods} & \multicolumn{2}{c|}{Metrics} & \multicolumn{3}{c}{User Preference} \\
\cline{2-3} \cline{4-6}
 & \small{CPBD$\uparrow$} & \small{ID$\uparrow$} & \small{Fide.$\uparrow$} & \small{Natur.$\uparrow$} & \small{Vis.Qua.$\uparrow$} \\
\hline
SadTalker & 0.3218 & 0.9188 & 4.15 & 3.12 & 3.97 \\
\hline
StyleHEAT & 0.2577 & 0.7993 & 3.26 & 3.65 & 3.7 \\
\hline
Ours & \textbf{0.4075} & \textbf{0.9293} & \textbf{4.8} & \textbf{3.97} & \textbf{4.52} \\
\bottomrule
\end{tabular}
}
\caption{Quantitative comparison and user study results for portrait image animation.}
\label{tab:metric_keypoint}
\end{minipage}
\vspace{-40pt}
\end{table*}

\subsection{Comparison with Other State-of-the-Art Methods}
\label{sec:compare}
In this paper, we propose a method that can perform various applications from different modalities by using a unified architecture. In this section, we demonstrate and compare our results with the state-of-the-art methods that specialize in each application field.

\noindent \textbf{Trajectory-based Image Animation.} 
For pure trajectory-based image animation, we compare our approach with DragNUWA~\cite{yin2023dragnuwa}. Their method integrates the sparse motion vector into the diffusion model without employing explicit flow and warping. Consequently, this makes it entirely uninterpretable and potentially causing various limitations. As depicted in the right part of Fig.~\ref{fig:comparison}, DragNUWA incorrectly extends the influence of the user annotation to the background, causing the beach to move to the right even though the user's intention was to control the ship's cruise direction. 
This issue is particularly prone to occur in DragNUWA when the motion prior to specific parts is overly strong.
In comparison, our method avoids this problem by explicitly determining the moving region through dense optical flow, thus yielding superior results. The quantitative results are shown on Tab.~\ref{tab:metric_traj}, where the proposed method shows much better performance in terms of all metrics.

Furthermore, our model can effectively control camera motion by employing various trajectory patterns by the suitable trajectories, including zooming in, zooming out, panning left, panning right, rotating clockwise, rotating counter-clockwise, \etc. We compare our approach with MotionCtrl~\cite{yin2023dragnuwa}. As shown in Fig.~\ref{fig:comparison}, our model achieves comparable results with MotionCtrl, with fewer visual artifacts and larger motion magnitudes.
Notably, since our method immediately uses the optical flow to control the video diffusion model, we can also control the camera motion directly using fixed optical flow patterns. Please refer to the supplementary for more details.

\begin{figure}[t]
\centering
\includegraphics[width=\linewidth]{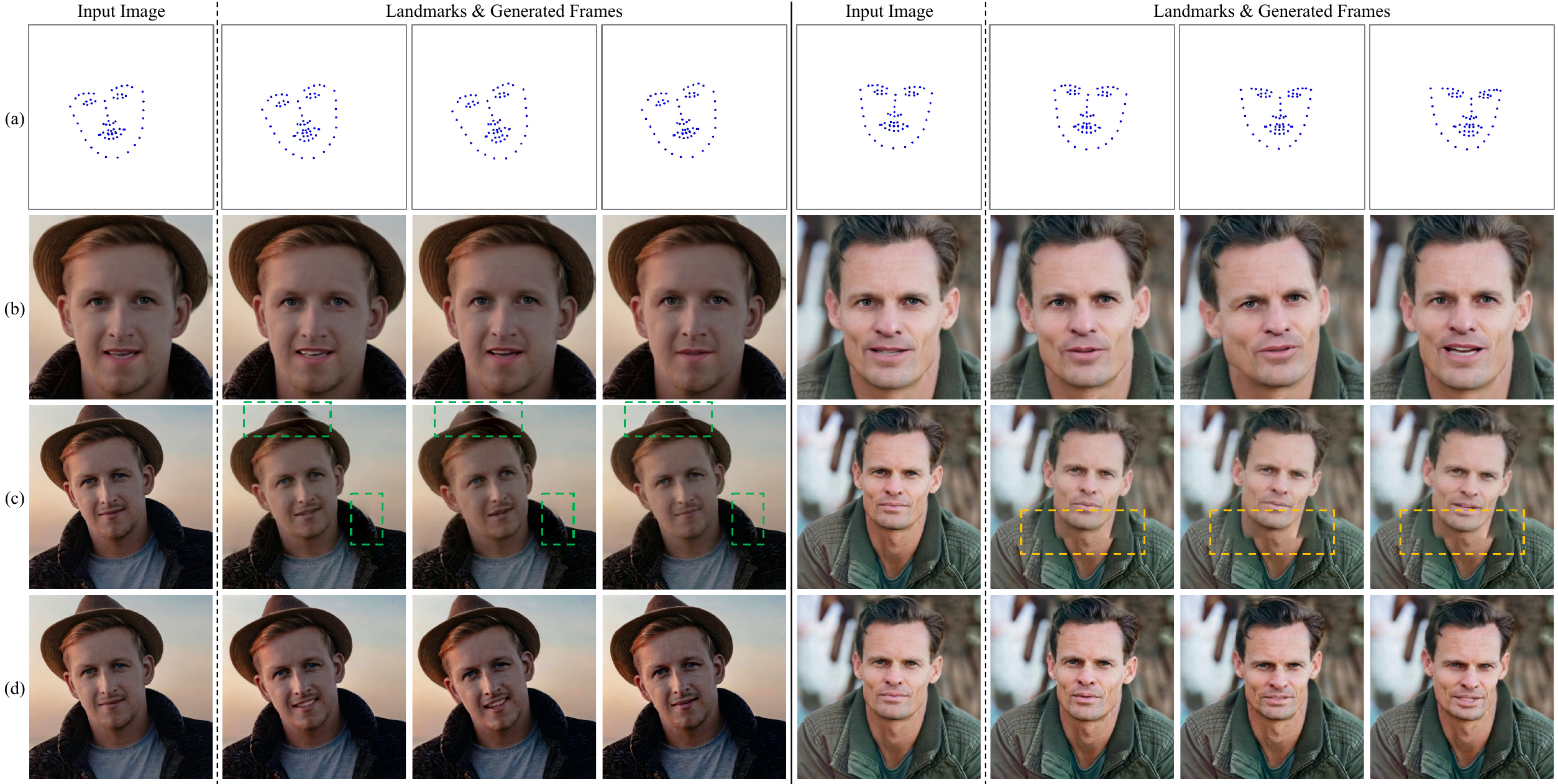}
\vspace{-2em}
\caption{
\textit{\textbf{Qualitative Comparisons for Portrait Animation Using Audio.}} (a)~Generated 3D Landmarks from SadTalker, (b)~StyleHEAT~\cite{yin2022styleheat}, (c)~SadTalker~\cite{zhang2023sadtalker}, (d)~Ours.
}
\label{fig:portrait_comparison}
\vspace{-10pt}
\end{figure}

\noindent \textbf{Portrait Image Animation.} We evaluate our keypoint-based MOFA-Adapter in comparison to prior state-of-the-art techniques such as StyleHEAT~\cite{yin2022styleheat}, Sad-Talker~\cite{zhang2023sadtalker}. The qualitative results are shown in Fig.~\ref{fig:portrait_comparison}. 
As StyleHEAT~\cite{yin2022styleheat} requires aligned images as input, we first crop and align the input image, then display the resulting output in the aligned format. 
We can see that StyleHEAT~\cite{yin2022styleheat} tends to generate overly smooth outcomes and struggles with identity preservation issues due to the inherent limitations of latent inversion via StyleGAN~\cite{karras2019style,karras2020analyzing}. SadTalker~\cite{zhang2023sadtalker} manages to process unaligned images using the paste-back technique, but it creates noticeable unnatural boundary artifacts. Additionally, it only controls head movement, leaving the target's body static during head motion, and may produce low-resolution outputs ($256 \times 256$), leading to a resolution discrepancy between facial and other parts. In comparison, our method effectively adapts to unaligned portrait images, removing noticeable paste-back artifacts, synchronizing body movements with head motion, and enabling high-resolution synthesis (up to $1024 \times 1024$) results. The corresponding quantitative results are reported in Tab.~\ref{tab:metric_keypoint}. Our method achieves the best results on all metrics, demonstrating its superiority over existing methods.

\noindent \textbf{User Studies.} We also conduct user studies to evaluate the effectiveness of the proposed method compared with baseline methods. For trajectory-based animation, we compare with DragNUWA using the same trajectories. For facial animation, we produce the same landmarks as SadTalker~\cite{zhang2023sadtalker} using the same audio. For each task, we choose 20 samples and invite 10 volunteers to evaluate the results. For trajectory-based animation, we consider the visual qualities and the control precision in the score range of 1 to 5~(higher the better). For facial animation, we consider the fidelity to the input image, the naturalness of the motion, and the visual qualities. The results are reported in Tab.~\ref{tab:metric_traj} and Tab.~\ref{tab:metric_keypoint}. We can see that the volunteers prefer the proposed method in terms of all the aspects.

\subsection{Ablation Studies}

\begin{figure}[t]
\centering
\includegraphics[width=\linewidth]{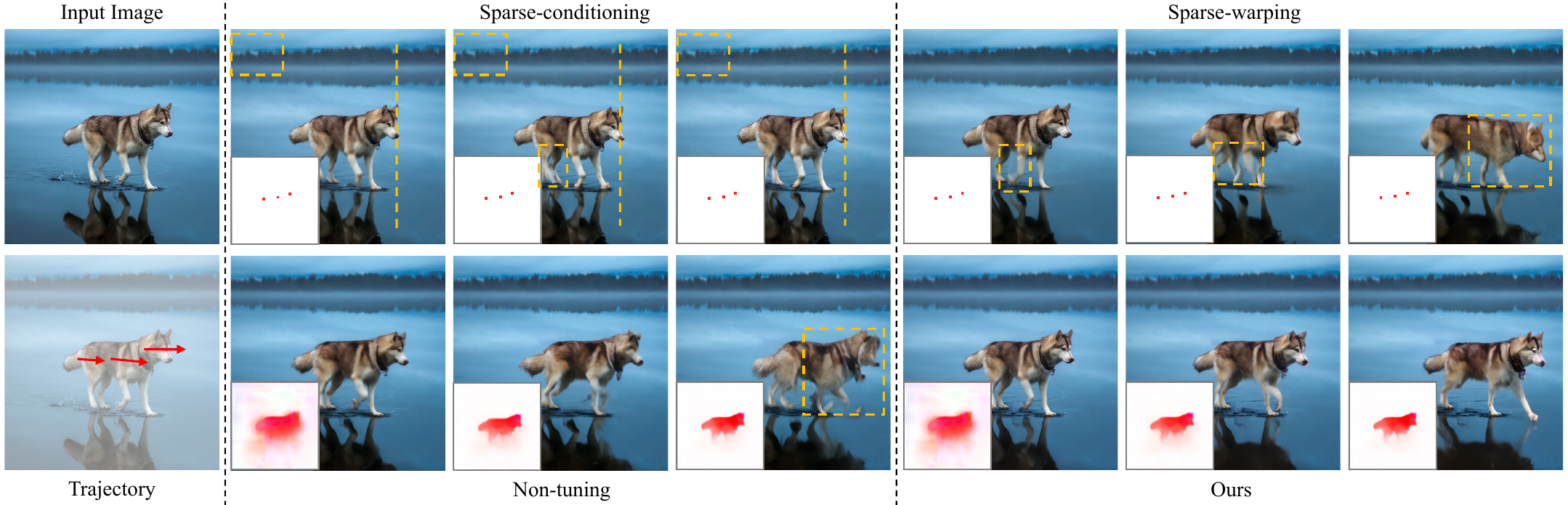}
\vspace{-2em}
\caption{
\textbf{\textit{Ablation on Network Design}.} Our full model achieves both the best controllability and synthesis quality compared to three baseline methods: 1) Sparse-conditioning model (w/o warping), 2) Sparse-warping model (w/o S2D network), and 3) Non-tuning model (w/o tuning).
}
\label{fig:ablation_traj}
\vspace{-10pt}
\end{figure}

\begin{figure}[t]
\centering
\includegraphics[width=\linewidth]{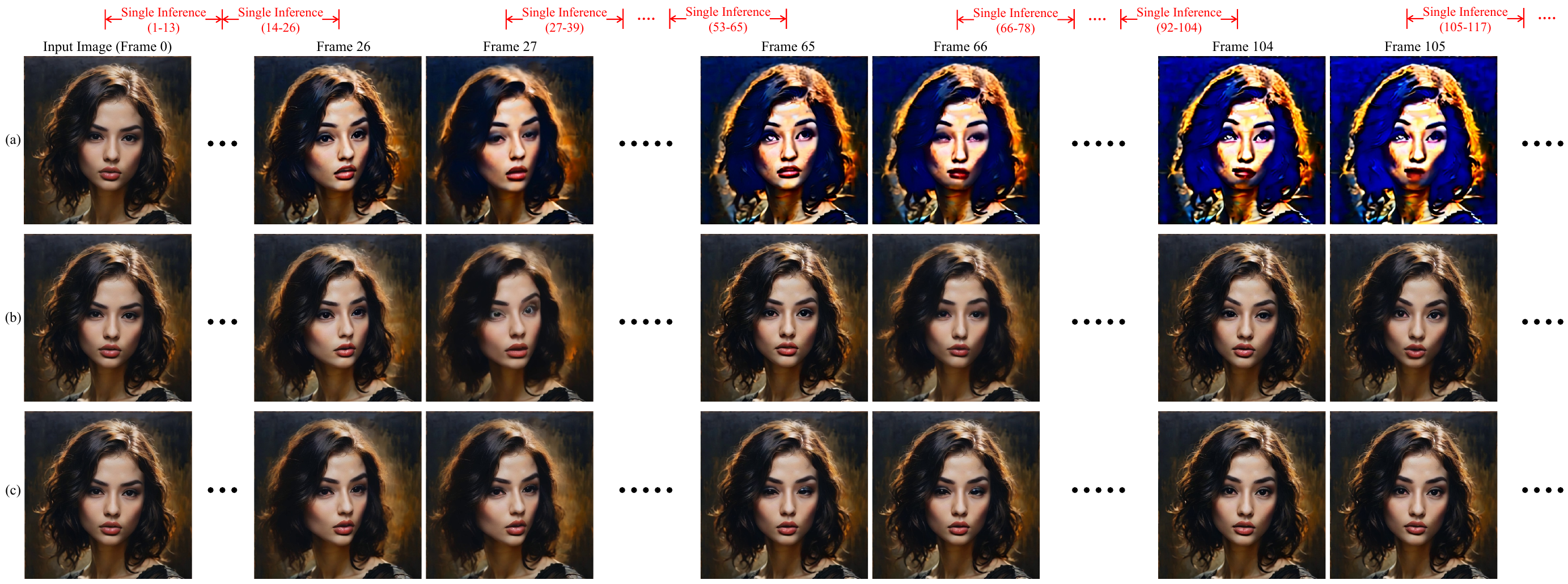}
\vspace{-2em}
\caption{
\textit{\textbf{Ablation on Longer Video Inference}}. (a)~Condition on the \textit{latest frame of previous inference} will show over-exposure results (Dynamic-conditional Naive Separation). (b)~Condition on the \textit{input frame and no frame overlap} will cause ``sudden changes'' between inferences clip (Zero-conditional Naive Separation). (c) Our proposed periodic sampling strategy can successfully generate the whole long sequences.
}
\label{fig:ablation_long}
\vspace{-10pt}
\end{figure}

\noindent \textbf{Network Structure of MOFA-Adapter.}
To evaluate the effectiveness of our network design, we compare the results of our method with three different variants:
\begin{itemize}
    \item \textbf{Sparse-conditioning model~(w/o warping).} We directly take the sparse trajectories with masks as conditions for Motion-aware ControlNet and do not use spatial warping in the feature space.
    \item \textbf{Sparse-warping model~(w/o S2D network).} We directly uses the sparse optical flow to spatially warp the deep features without using the proposed S2D model to turn it into a dense flow representation.
    \item \textbf{Non-tuning model~(w/o tuning).} Directly use the flow-based reconstruction model, without further finetuning.
\end{itemize}

The corresponding results are in Fig.~\ref{fig:ablation_traj}. The sparse-conditioning model is unable to precisely manipulate the target object's trajectory, as the dog does not move to the right, and the background slightly shifts to the left. We argue that this issue arises from the spatial misalignment caused by not employing the spatial warping operation.
While the sparse-warping model can successfully manage the target's trajectory, it struggles to effectively transform the deep feature from the initial frame due to the lack of dense feature control provided by dense optical flow, resulting in significant artifacts in the output. 
The non-tuning model, however, experiences over-control from the dense optical flow and is unable to generate suitable content using the strong prior of the SVD.
In comparison, our comprehensive model achieves optimal results.

\noindent \textbf{Longer Video Inference.} Additionally, we evaluate our periodic sampling approach against two alternative strategies:
\begin{itemize}
    \item \textbf{Dynamic-conditional Naive Separation} separates all frames into groups of 13, generating each group based on the \textit{generated last frame of the preceding group}, and then combines all the frames.
    \item \textbf{Zero-conditional Naive Separation} separates all frames into groups of 13, generating each group based on the \textit{first frame}, and then combines all the frames.
\end{itemize}
As shown in Fig.~\ref{fig:ablation_long}, the Dynamic-conditional Naive Separation strategy~(a) leads to significant error accumulation, resulting in unsatisfactory results in the later stages of the long generation process. Although the Zero-conditional Naive Separation strategy~(b) mitigates error accumulation by consistently conditioning on the first frame, noticeable flickering effects and temporal inconsistency are apparent between the first frame of the current group and the last frame of the previous group, such as frames 26 and 27, 65 and 66, and 104 and 105 in Fig.~\ref{fig:ablation_long}. In comparison, our periodic sampling algorithm~(c) effectively addresses both error accumulation and temporal inconsistency by blending features from neighboring groups in the latent space. Please refer to the supplementary video results, which provide a more intuitive and clear demonstration of the superiority of periodic sampling.

\begin{figure}[t]
\centering
\includegraphics[width=\linewidth]{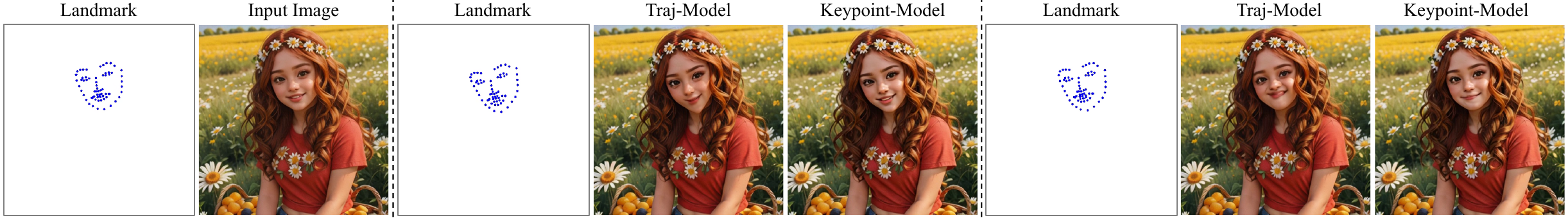}
\vspace{-2em}
\caption{
\textit{\textbf{Ablation on Domain-Aware MOFA-Adapter}}. Using the model in the trajectory domain without specific tuning for portrait animation will lead to unsatisfactory results due to the domain gap.
}
\label{fig:ablation_portrait}
\vspace{-15pt}
\end{figure}

\noindent \textbf{The effectiveness of Domain-Aware MOFA-Adapter.}
As introduced in Sec.~\ref{sec:method}, we train different MOFA-Adapters for different motion domains. To evaluate the effectiveness and the necessity of domain-specific tuning, we directly use our trajectory-based model to perform landmark-based portrait image animation. As shown in Fig.~\ref{fig:ablation_portrait}, directly applying the model trained on open-domain will cause unnatural facial expressions.

\subsection{Limitations}

Although our method achieves remarkable progress in controllable image animation, it still has some limitations. First, different from SORA~\cite{sora}, our method is hard to control~(generate) new content which is far from the given image since the current video diffusion model only trains on limited video clips. Second, our model may suffer from visual artifacts including blurriness or structure loss under large motion guidance. Visual examples are presented in the supplementary.
\section{Conclusion}

In this paper, we introduce a novel pipeline for generic and controllable image-to-video animation from multiple motion domains~(\eg, handcrafted trajectories, dense flows, and human key-points). To achieve this goal, we design the sparse-to-dense MOFA-Adapter to control the generated motions in the video generation pipeline. Powered by the proposed framework, we can achieve controllable video generation via fine-grained control which is unified as flow motion fields. The experiments show the advantage of the proposed framework over current state-of-the-art methods for various applications.

\appendix

\section{Implementation Details}
\addcontentsline{toc}{chapter}{Implementation Details}

\subsection{More Architecture Details of MOFA-Adapter}

\begin{figure}[t]
\centering
\includegraphics[width=0.7\linewidth]{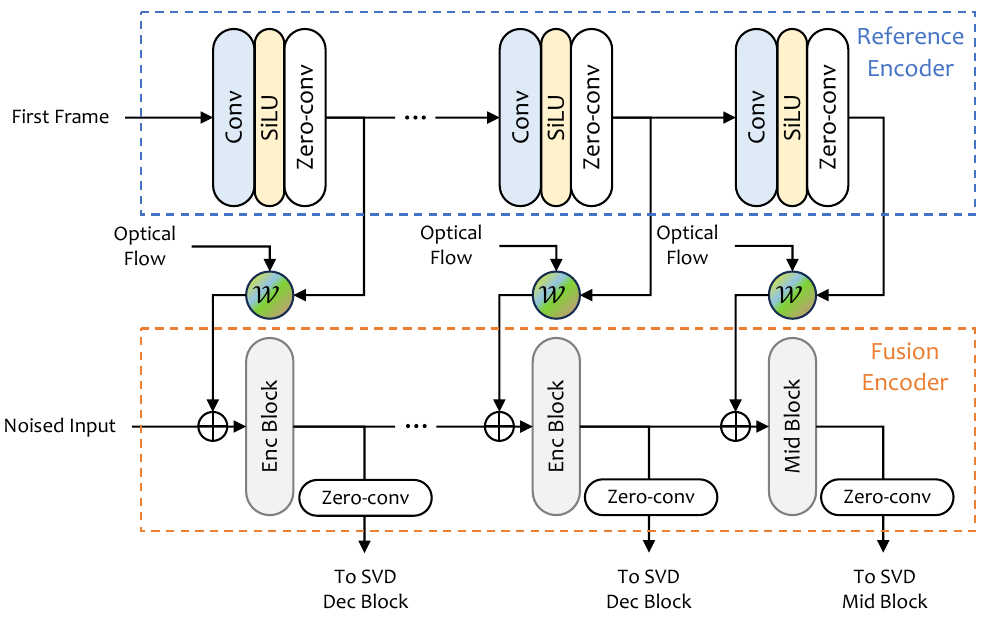}
\caption{
\textit{\textbf{Detailed architecture of MOFA-Adapter.}}
}
\label{fig:architecture}
\end{figure}

The proposed MOFA-Adapter is composed of three components: 1) Sparse-to-Dense Motion Generation Network (S2D network), 2) Reference Encoder, and 3) Fusion Encoder. We show the detailed architecture for feature merging in Fig.~\ref{fig:architecture}. The Fusion Encoder's architecture is identical to that of the SVD~\cite{svd} Encoder. Forward warping is utilized for spatial warping operations within the feature space.

\subsection{More Training Details}

The trajectory-based model is trained on the WebVid-10M dataset ~\cite{webvid} using the AdamW optimizer with a learning rate of $2 \times 10^{-5}$. The batch size is set at 8, and the total number of training iterations is 100,000. The portrait-based model is trained on a self-compiled dataset that includes 5,889 different human portrait videos. The AdamW optimizer is used, with a learning rate of $2 \times 10^{-5}$. The batch size is set to 1, and the total training iteration is 200,000.

\subsection{Inference via Multiple MOFA-Adapters}

As indicated in the main paper, we can integrate multiple MOFA-Adapters for more sophisticated and complex control using control signals from various modalities. For instance, users can merge the landmark signal with handcrafted trajectories. Specifically, we first route the trajectory control signals and landmark signals through the MOFA-Adapter for each modality separately. Contrary to the original Multi-ControlNet~\cite{controlnet} algorithm, we employ a mask-aware strategy to define the control area for each MOFA-Adapter. Specifically, the user can designate the region where the landmark signal is accountable for, such as the human face region. Based on this mask, we extract the deep feature of the multiscale output of the landmark-based MOFA-Adapters within the mask region, and that of the trajectory-based MOFA-Adapters outside the mask region. Finally, we input the combined features into the frozen SVD to obtain the final output.

\section{More Visual Results}
\addcontentsline{toc}{chapter}{More Visual Results}

In this section, we demonstrate more results generated by our methods. For video results, please refer to the video demo provided in the supplementary.

\subsection{Trajectory-based Image Animation}

\begin{figure}[t]
\centering
\includegraphics[width=\linewidth]{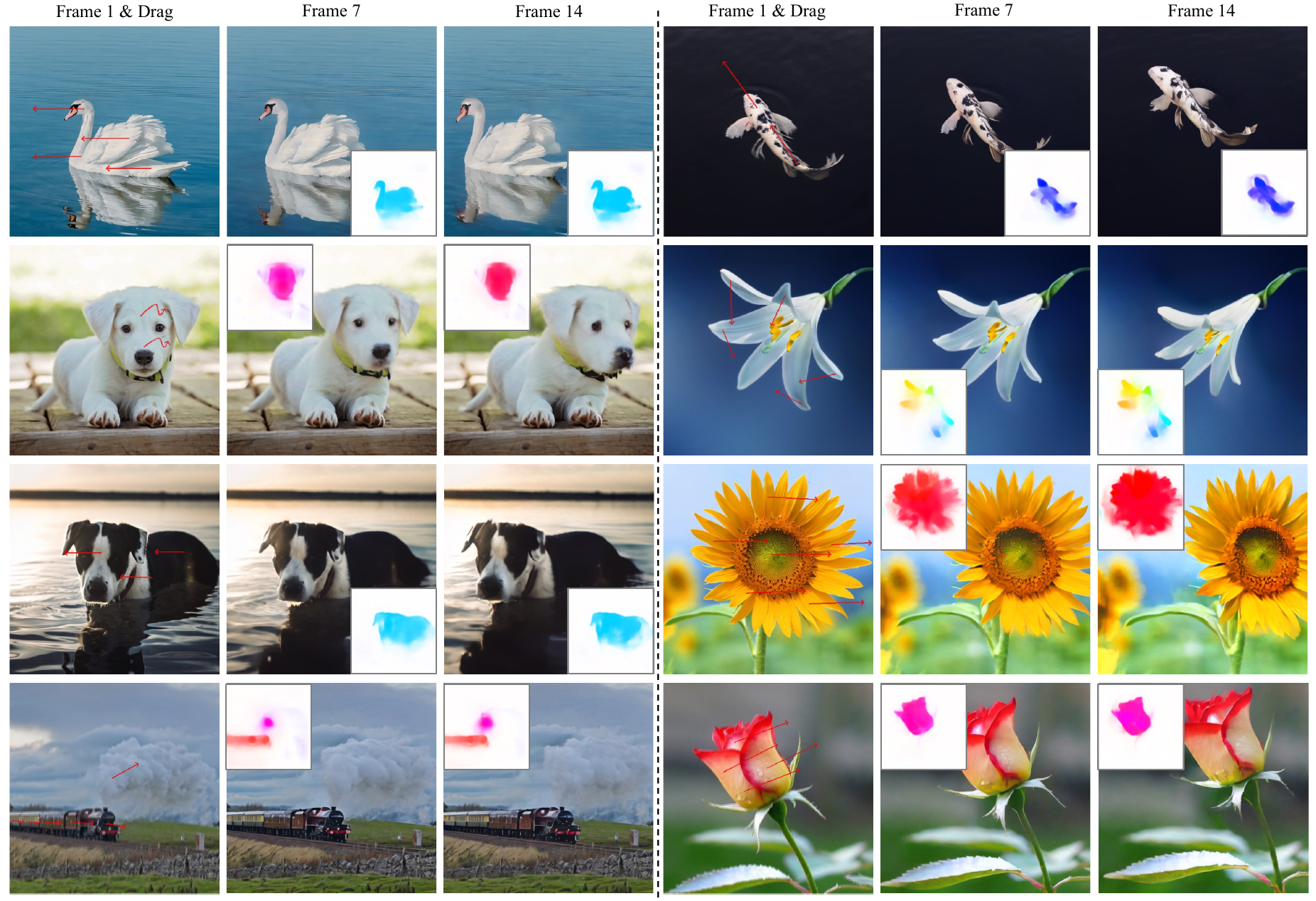}
\caption{
\textit{\textbf{More visual results for trajectory-based image animation.} }
}
\label{fig:trajectory}
\end{figure}

The Trajectory-based Image Animation results are demonstrated in Fig.~\ref{fig:trajectory}.

\subsubsection{Camera Motion Control}

\begin{figure}[t]
\centering
\includegraphics[width=\linewidth]{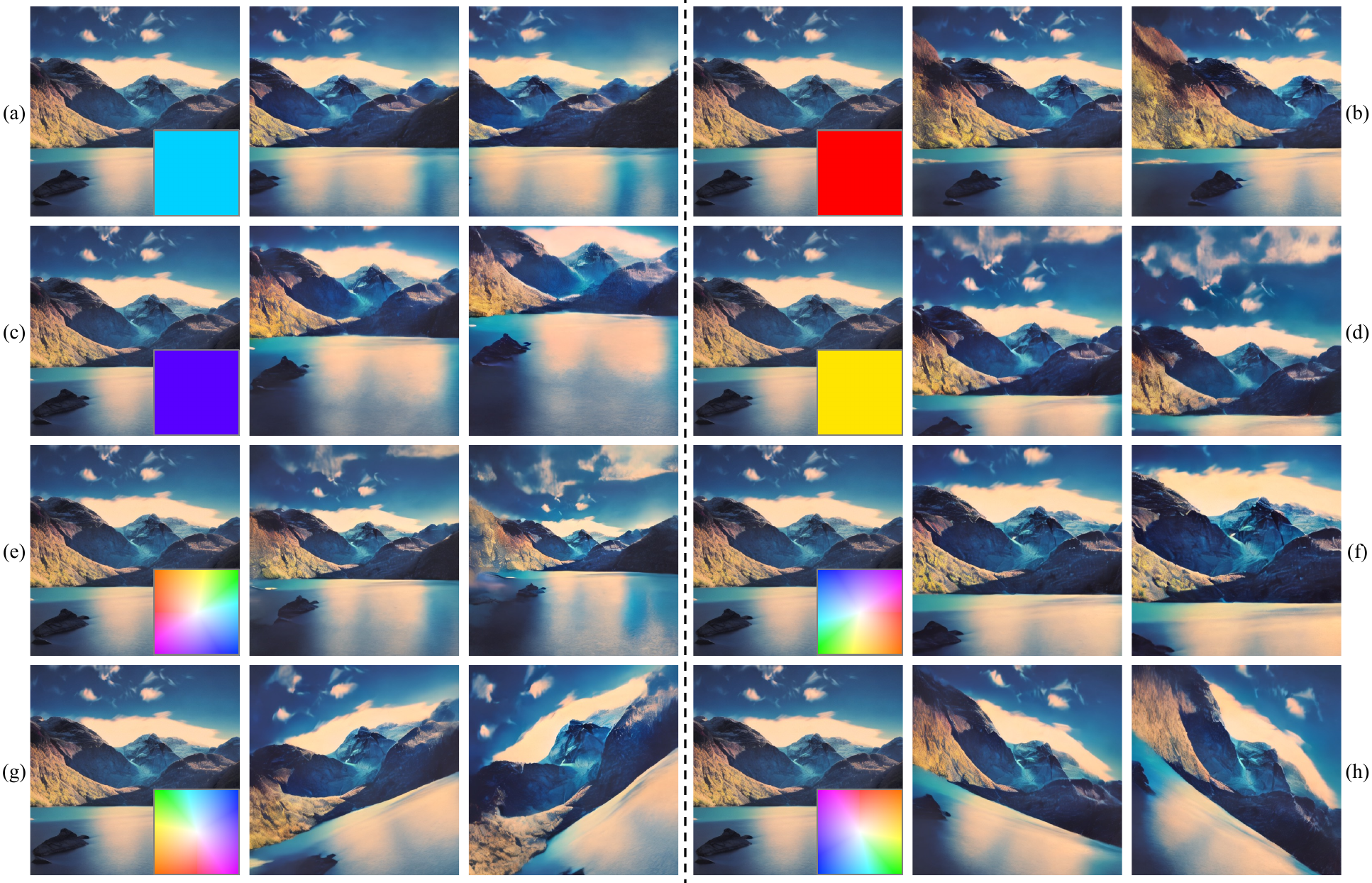}
\caption{
\textit{\textbf{Camera motion control via fixed optical flow patterns.}} (a)~Pan Right, (b)~Pan Left, (c)~Pan Down, (d)~Pan Up, (e)~Zoom Out, (f)~Zoom In, (g)~Clockwise, (h)~Counter-Clockwise. 
}
\label{fig:camera_motion_fixflow}
\end{figure}

As stated in the main paper, besides handcrafted trajectories, our model is also capable of controlling camera motion via basic optical flow patterns. The corresponding results are illustrated in Fig.~\ref{fig:camera_motion_fixflow}.

\subsection{Portrait Image Animation}

\begin{figure}[t]
\centering
\includegraphics[width=\linewidth]{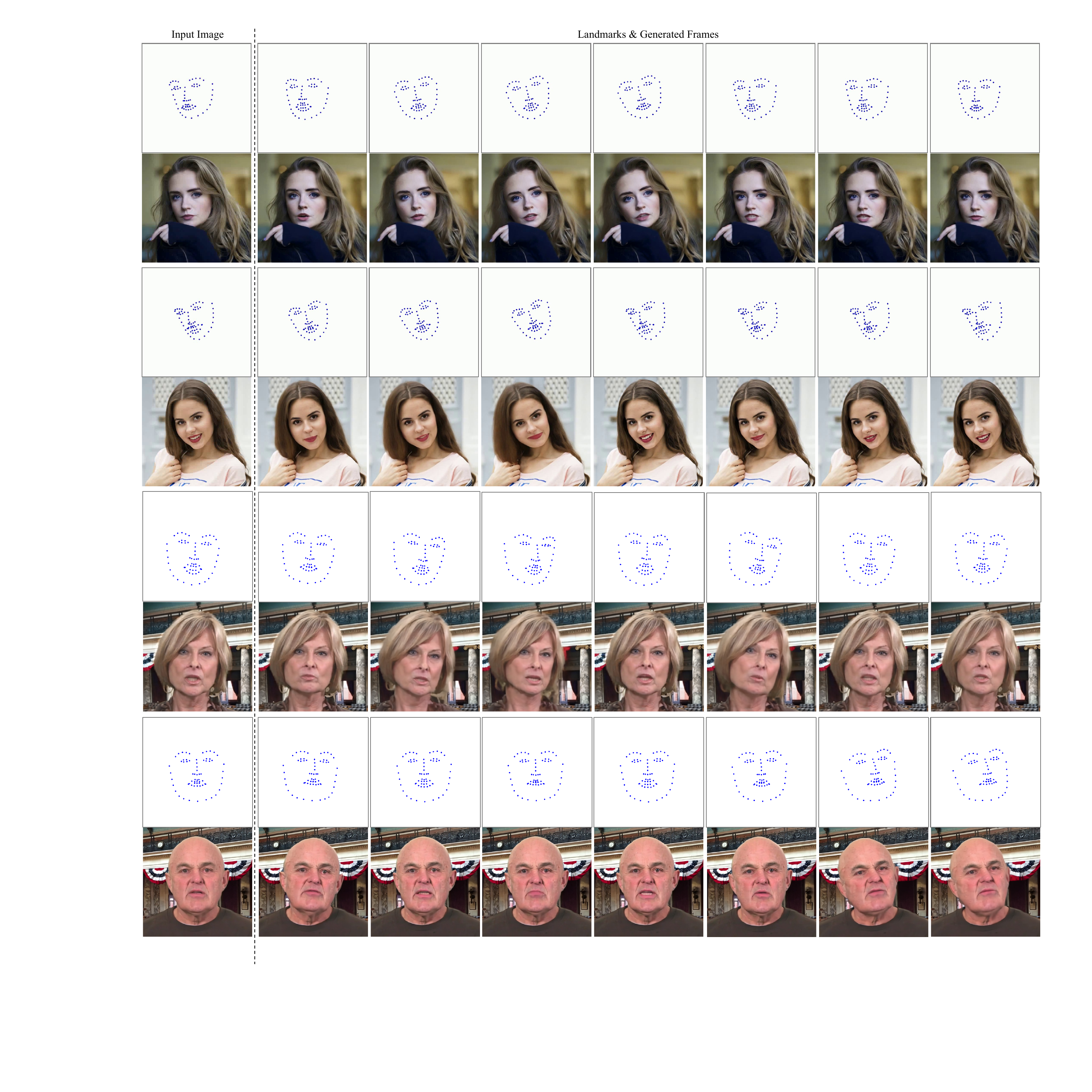}
\caption{
\textit{\textbf{More visual results for portrait image animation.}}
}
\label{fig:portrait}
\end{figure}

More portrait image animation results are demonstrated in Fig.~\ref{fig:portrait}.

\subsection{Multi-MOFA Adapters}

\begin{figure}[t]
\centering
\includegraphics[width=\linewidth]{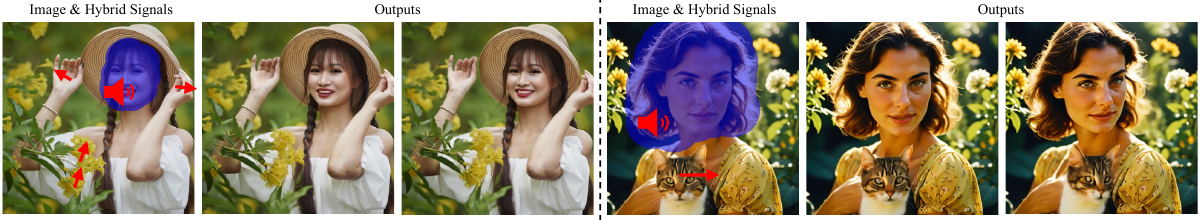}
\caption{
\textit{\textbf{Visual results for advanced control with Multi-MOFA Adapters.}}
}
\label{fig:hybrid}
\end{figure}

More advanced control results via Multi-MOFA Adapters are demonstrated in Fig.~\ref{fig:hybrid}.

\section{Comparison Results}
\addcontentsline{toc}{chapter}{Comparison Results}

In this section, we demonstrate more comparison results against other methods. For video results, please refer to the video demo provided in the supplementary.

\subsection{Trajectory-based}

\begin{figure}[t]
\centering
\includegraphics[width=\linewidth]{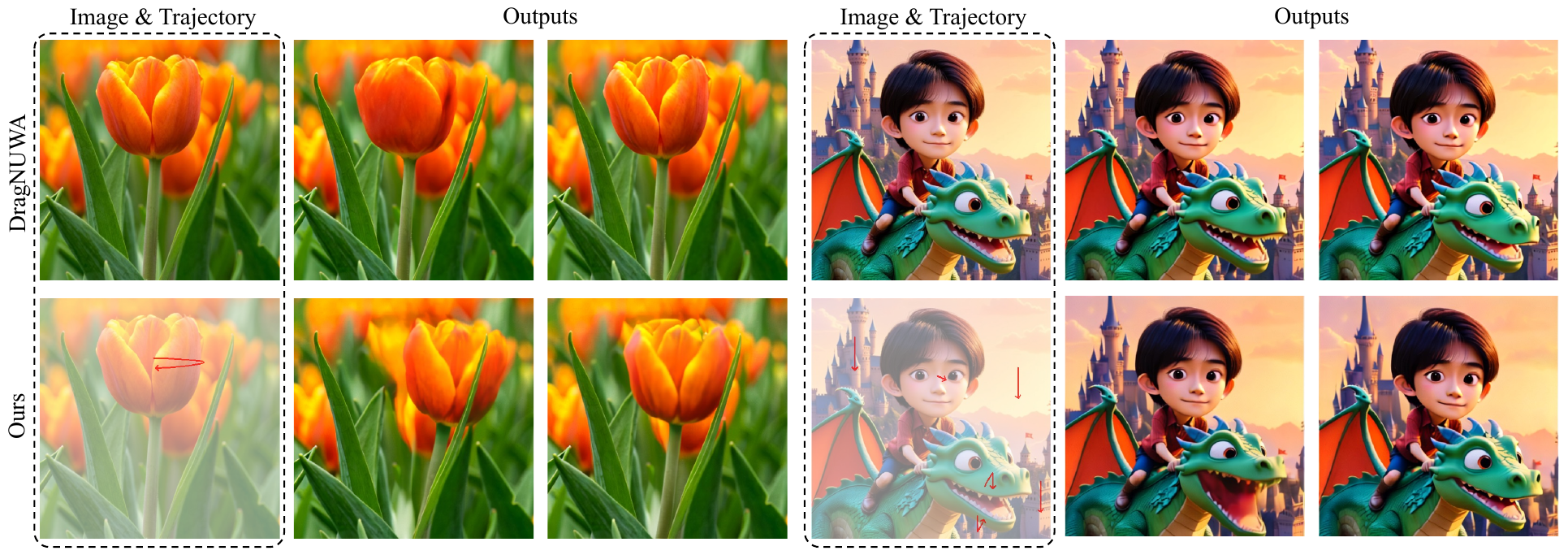}
\caption{
\textit{\textbf{Visual comparisons with DragNUWA~\cite{yin2023dragnuwa} for trajectory-based image animation.}}
}
\label{fig:traj_compare}
\end{figure}

More comparison results with DragNUWA~\cite{yin2023dragnuwa} are demonstrated in Fig.~\ref{fig:traj_compare}.

\subsubsection{Motion Brush}

\begin{figure}[t]
\centering
\includegraphics[width=\linewidth]{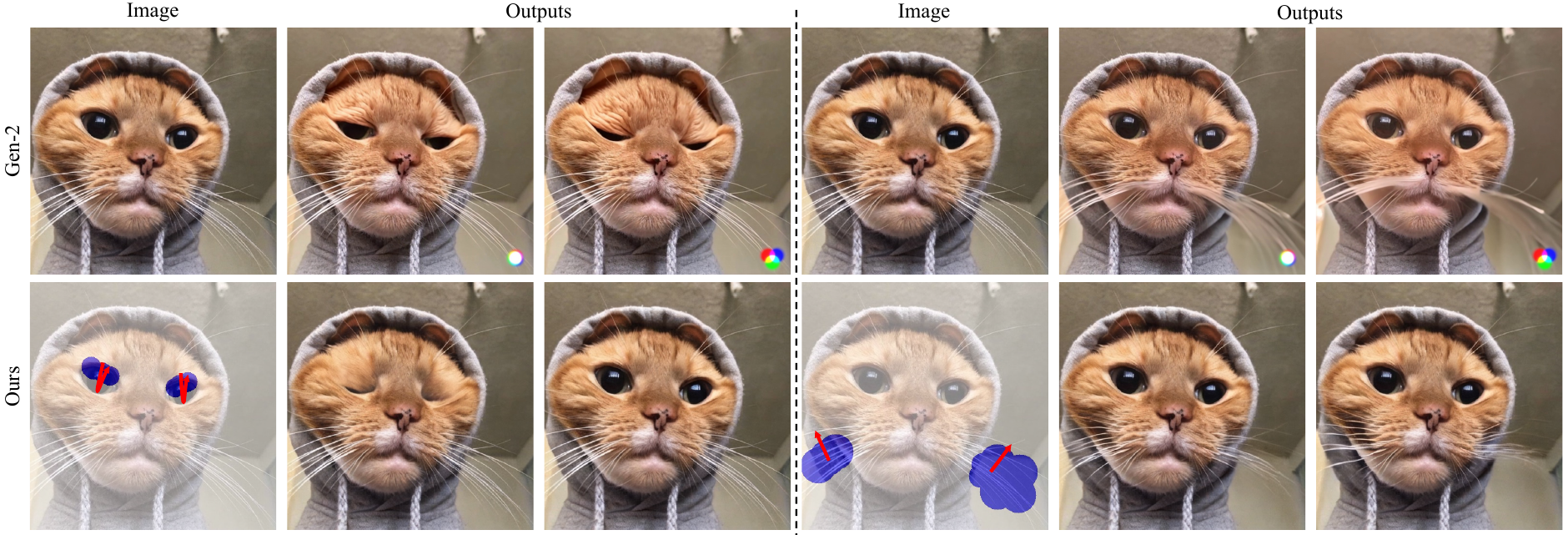}
\caption{
\textit{\textbf{Image animation results from our method and Gen-2~\cite{gen2}.}} Gen-2 employs a mask + direction approach, which is not suitable for managing complex motions. In contrast, our method integrates trajectory control with motion brushes, enabling advanced non-linear control (\eg, blinking) for the target objects.
}
\label{fig:motion_brush}
\end{figure}

As stated in the main paper, we can employ motion mask brushes to attain detailed control by designating the spatial region of the flow patterns since our method utilizes intermediate optical flow patterns for motion control. Gen-2~\cite{gen2} also supports motion brushes, but it only supports basic directions (Mask + direction) and is incapable of executing non-linear complex controls (for instance, blinking). Our method combines regional mask and trajectory (mask + trajectory), being able to handle advanced nonlinear motions. The comparative results corresponding to this are displayed in Fig.~\ref{fig:motion_brush}.

\subsection{Portrait Image Animation}

\begin{figure}[t]
\centering
\includegraphics[width=\linewidth]{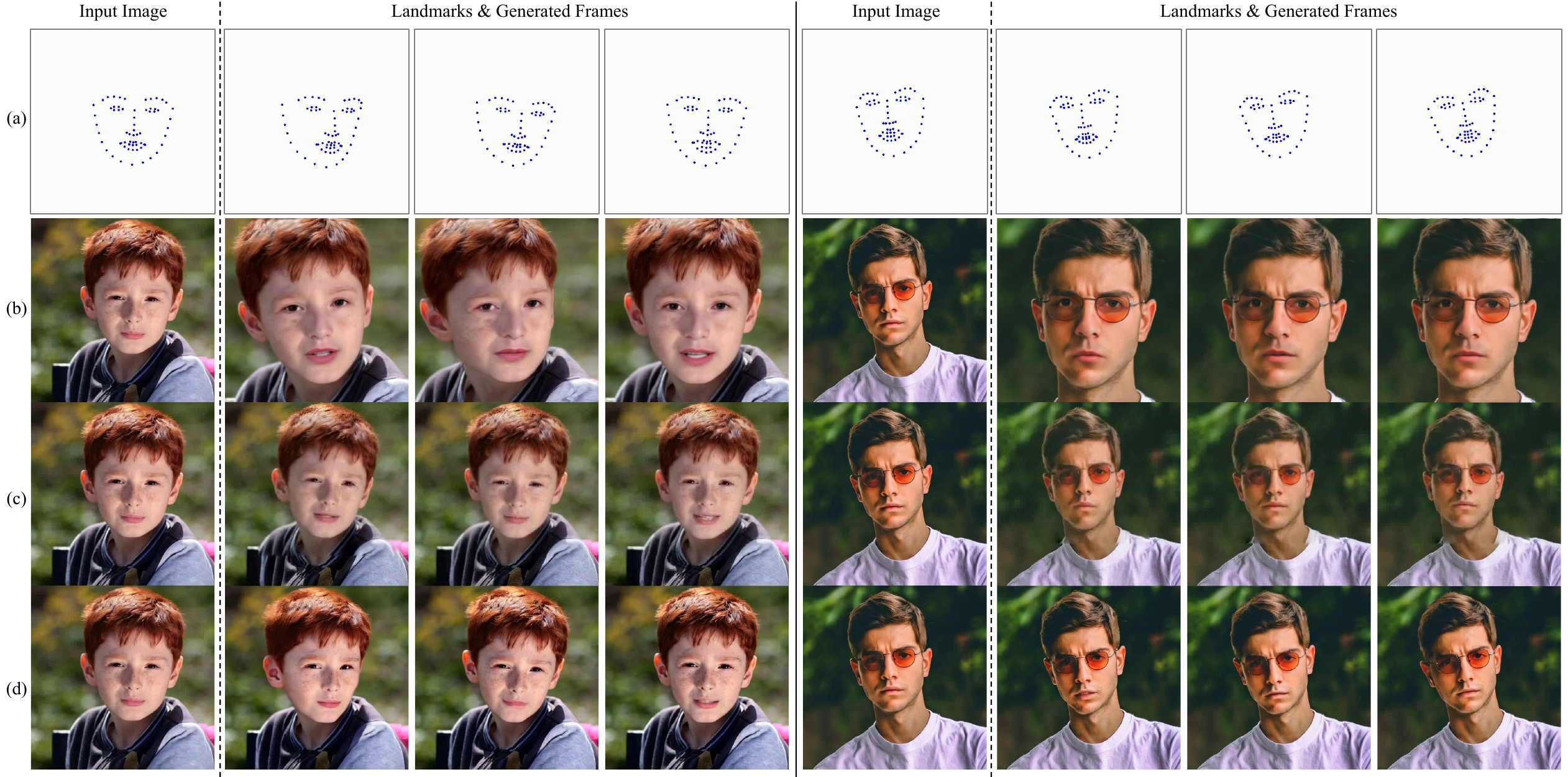}
\caption{
\textit{\textbf{More visual comparisons for portrait image animation.}} (a) StyleHEAT~\cite{yin2022styleheat}, (b) Sadtalker~\cite{zhang2023sadtalker}, (c) Ours.
}
\label{fig:portrait_comparison_supp}
\end{figure}

More visual comparison results with StyleHEAT~\cite{yin2022styleheat} and SadTalker~\cite{zhang2023sadtalker} are demonstrated in Fig.~\ref{fig:portrait_comparison_supp}. We also give more quantitative results with our methods and SadTalker~\cite{zhang2023sadtalker} on visual quality~(LPIPS). The proposed method shows a much better performance on visual quality~(0.2099) than SadTalker~(0.2308). In addition, our method also shows comparable results in lip synchronization. We give some examples in the supplementary video.

\section{Ablation Study Results}
\addcontentsline{toc}{chapter}{Ablation Study Results}

We also consider the quantitative comparison of ablation studies in network structure. The same dataset is used for evaluation as the one used for quantitative comparisons with DragNUWA~\cite{yin2023dragnuwa} in the main paper. 
As shown in Tab.~\ref{tab:metric_ablation_traj}, the proposed full method achieves the most balanced results in terms of all metrics. Our method w/o tuning of the MOFA-Adapter shows very limited motions compared with our full methods. Our method w/o S2D shows second-best results. However, from LPIPS, the generated video is different from the motion guidance. Finally, our method uses explicit warping as motion control, removing the explicit motion warping shows much worse results in all metrics.
For video results, please refer to the video demo provided in the supplementary. We also provide the video ablation results for longer video generation and domain-aware MOFA-Adapter in the video demo.

\section{Video Demo}
\addcontentsline{toc}{chapter}{Video Demo}

We provide the video demo in the supplementary, which includes brief introduction, video results, ablation studies, and the limitations of our method.

\section{Limitations}
\addcontentsline{toc}{chapter}{Limitations}

\begin{figure}[t]
\centering
\includegraphics[width=\linewidth]{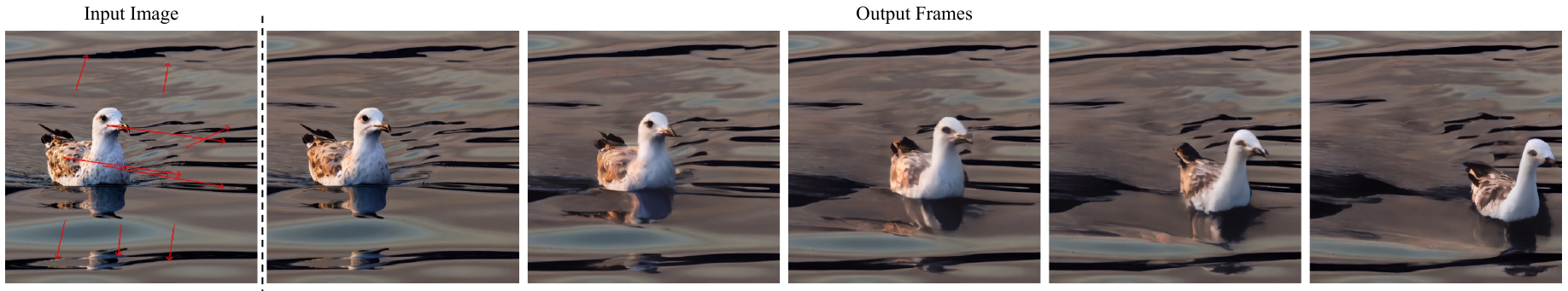}
\caption{
\textit{\textbf{Limitation of our method.}} Our model may encounter visual artifacts such as loss of structure or blurriness under extensive motion guidance.
}
\label{fig:limitation}
\end{figure}

Unlike SORA~\cite{sora}, our method struggles to control or generate new content that is significantly different from the provided image, as the current video diffusion model is only trained on a limited number of video clips. Additionally, our model may encounter visual artifacts such as blurriness or loss of structure under extensive motion guidance. Visual examples of these issues are provided in Fig.~\ref{fig:limitation}.

\begin{table*}[t] 
\centering
\begin{minipage}[b]{1\textwidth}
\centering
\begin{tabular}{l|cccc}
\toprule
Methods & \small{LPIPS $\downarrow$} & \small{FID $\downarrow$} & \small{FVD $\downarrow$} \\
\hline
w/o warping  & 0.2619 & 18.80 & 184.27 \\
\hline
w/o S2D & 0.2376 & \underline{16.87} & \textbf{81.80} \\
\hline
w/o tuning & \textbf{0.2163} & 16.97 & 102.17 \\
\hline
Ours & \underline{0.2274} & \textbf{16.82} & \underline{86.76} \\
\bottomrule
\end{tabular}
\caption{Quantitative comparison results for ablation study on trajectory-based image animation.}
\label{tab:metric_ablation_traj}
\end{minipage}
\end{table*}


\bibliographystyle{splncs04}
\bibliography{main}
\end{document}